\documentclass[lettersize,journal]{IEEEtran}
\usepackage{amsmath,amsfonts}
\usepackage{algorithmic}
\usepackage{algorithm}
\usepackage{array}
\usepackage[caption=false,font=normalsize,labelfont=sf,textfont=sf]{subfig}
\usepackage{textcomp}
\usepackage{stfloats}
\usepackage{url}
\usepackage{verbatim}
\usepackage{graphicx}
\usepackage{cite}
\usepackage{multirow}
\usepackage{color}
\usepackage{booktabs}
\begin{document}

\title{MP-MVS: Multi-Scale Windows PatchMatch\\ and Planar Prior Multi-View Stereo}

\author{Rongxuan Tan, Qing Wang, Xueyan Wang, Chao Yan, Yang Sun and Youyang Feng
        % <-this % stops a space
\thanks{This paper was produced by the School of Instrument Science and Engineering, Southeast University. They are in Nanjing, China.}
}


\maketitle

\begin{abstract}
Significant strides have been made in enhancing the accuracy of Multi-View Stereo (MVS)-based 3D reconstruction. However, untextured areas with unstable photometric consistency often remain incompletely reconstructed. In this paper, we propose a resilient and effective multi-view stereo approach (MP-MVS). We design a multi-scale windows PatchMatch (mPM) to obtain reliable depth of untextured areas. In contrast with other multi-scale approaches, which is faster and can be easily extended to PatchMatch-based MVS approaches. Subsequently, we improve the existing checkerboard sampling schemes by limiting our sampling to distant regions, which can effectively improve the efficiency of spatial propagation while mitigating outlier generation. Finally, we introduce and improve planar prior assisted PatchMatch of ACMP. Instead of relying on photometric consistency, we utilize geometric consistency information between multi-views to select reliable triangulated vertices. This strategy can obtain a more accurate planar prior model to rectify photometric consistency measurements. Our approach has been tested on the ETH3D High-res multi-view benchmark with several state-of-the-art approaches. The results demonstrate that our approach can reach the state-of-the-art. The associated codes will be accessible at https://github.com/RongxuanTan/MP-MVS.
\end{abstract}


\section{Introduction}
\IEEEPARstart{M}{ulti-View Stereo}  (MVS) has been a hot problem for research in computer vision. Thanks to the success of the Structure-from-Motion (SfM) algorithm \cite{ref1,ref2,ref3} and the availability of public benchmarks \cite{ref4,ref5,ref6,ref7}, many excellent MVS approaches have emerged in the last decade, and these works have achieved impressive results. Presently, PatchMatch-based approaches are still the best way to solve the dense matching problem efficiently and reliably. PatchMatch estimates the plane of the object surface corresponding to each pixel, thereby transforming depth estimation into plane estimation. The initial plane hypotheses of pixels are first randomly initialized. Then, the confidence of these plane hypotheses is computed based on photometric consistency. Finally, reliable plane hypotheses are propagated to neighboring pixels and the depth of each pixel is estimated efficiently.

Currently, some of the most advanced PatchMatch-based MVS approaches can acquire accurate point clouds. However, PatchMatch \cite{ref8} is profoundly reliant on the outcomes of the photometric consistency metric. When the patch is located in untextured areas, there is a lack of sufficient texture information in the receptive field, leading to erroneous depth estimates, which eventually makes it difficult to reconstruct these scenes completely. High-resolution images tend to contain more untextured areas, which further exacerbates this issue. Therefore, it is still a challenging problem to accurately estimate the depth of untextured areas.

The texture information within a patch becomes more abundant when an image is downsampled. Therefore ACMM\cite{ref9} employs image downsampling and obtains more reliable depth estimation. Several of the most advanced PatchMatch-based MVS approaches\cite{ref10,ref11,ref12} use ACMM as a fundamental module. Similarly, MARMVS\cite{ref13} ensures that there is enough texture information within the receptive field by adjusting the patch size of each pixel. These strategies exhibit enhanced robustness in handling scenes with varying scales and texture richness.

In this paper, we also present a multi-scale depth estimation approach to improve the accuracy and completeness of depth estimation in untextured areas, and the contribution of our proposed approach is as follows:

\begin{itemize}
\item{We propose an efficient and straightforward multi-scale windows PatchMatch (mPM). It conducts photometric consistency using multi-scale patches in sequence within a single PatchMatch process. Notably, it outperforms ACMM in terms of speed and can be simply integrated into any existing PatchMatch-based MVS approaches.}
\item{We have refined the existing checkerboard sampling schemes by limiting our sampling to eight distant regions from the pixels. This improvement effectively curbs the generation of outliers, enhancing both the efficiency and reliability of correct hypothesis propagation.}
\item{We have adopted and improved the planar prior assisted PatchMatch of ACMP\cite{ACMP}. Reliable triangulated vertices are selected by geometric consistency information between multiple views instead of photometric consistency, which improves the accuracy of the planar prior model.}
\end{itemize}

Finally, we combine the above modules to implement the PatchMatch-based MVS approach (MP-MVS). We tested our approach on the ETH3D high-resolution multi-view benchmark. The results show that our approach reaches the state-of-the-art.

\section{RELATED WORK}
MVS can be divided into four classes according to the scene representation: voxel-based \cite{ref14,ref15,ref16}, surface evolution-based \cite{ref17,ref18,ref19}, feature growing-based \cite{ref20,ref21,ref22}, depth map merging-based. Our approach belongs to the last class. In this section, we only introduce PatchMatch-based related approaches. 

\begin{figure*}[!t]\centering\includegraphics[width=18cm]{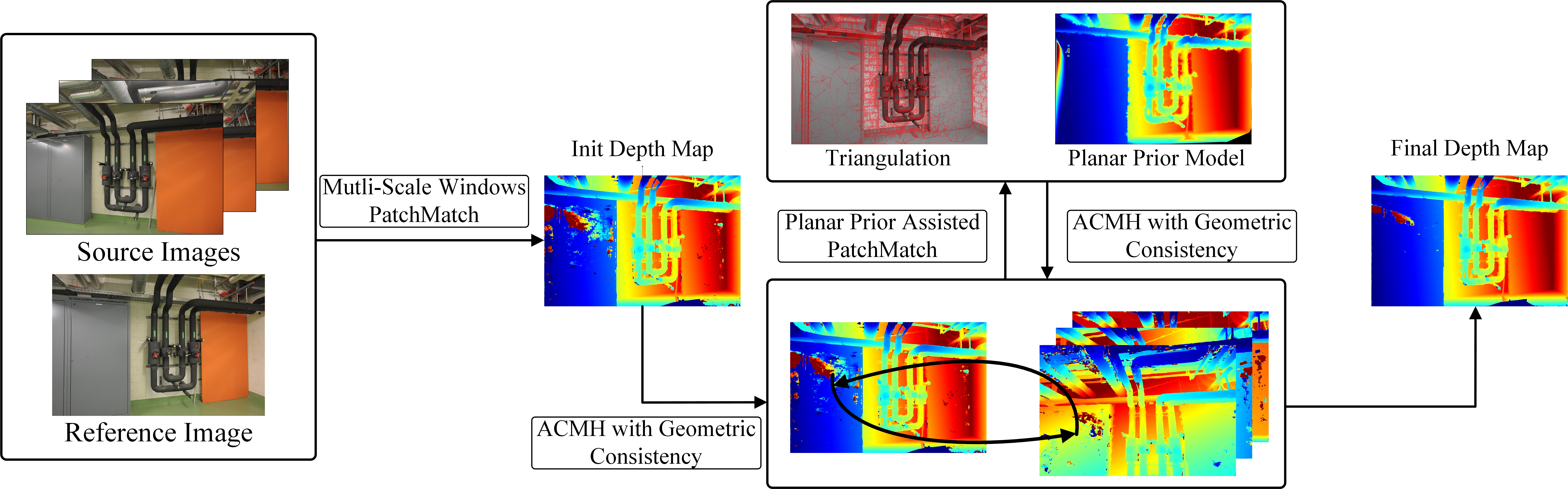}\caption{Overview of the proposed MP-MVS. The initial depth map is first computed using a multi-scale windows PathMatch, and then the initial depth map is optimized for geometric consistency and planar prior assisted PatchMatch. Finally, we perform a geometric consistency optimization to obtain the final depth map.}\label{fig_1}
\end{figure*}

\begin{figure}[!t]\centering\includegraphics[width=8.5cm]{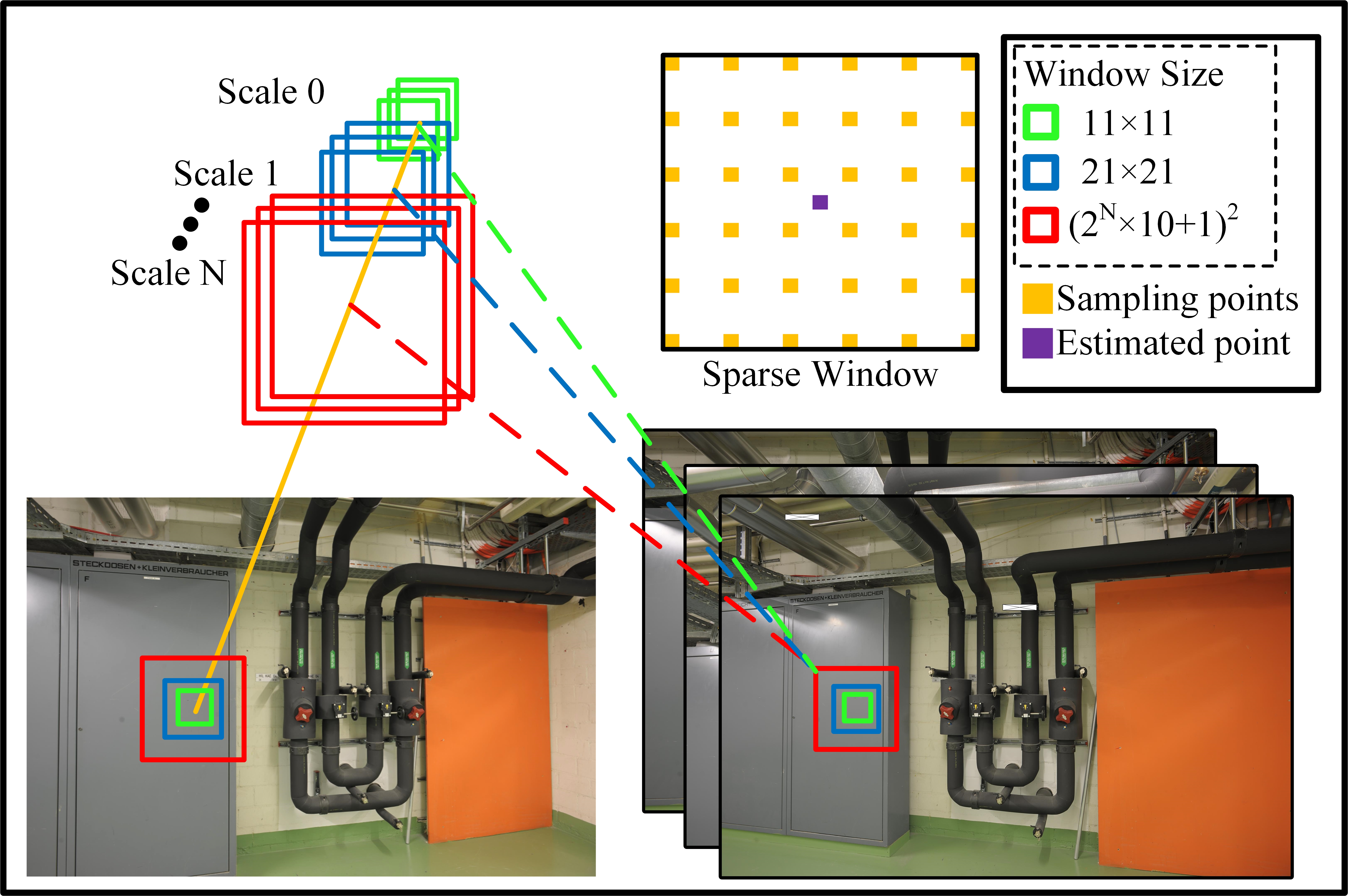}\caption{Mutli-scale windows PatchMatch. The sparse window shows how our multi-scale window is sparsely sampled.}\label{fig_2}
\end{figure}

Gipuma \cite{ref23} designed a diffusion-like propagation scheme that enables the implementation of PatchMatch on GPUs. COLMAP \cite{ref24} improves the robustness of view selection through photometric and geometric prior, and it incorporates geometric consistency optimization into the PatchMatch inference process to refine the depth map. Although the above approaches can achieve impressive accuracy, these approaches have difficulty in achieving reliable depth estimation for untextured areas and non-lambert surfaces. To deal with this problem, some approaches estimate the depth of untextured areas by expanding the receptive field of the patch. ACMM and PLC\cite{ref25} estimate a reliable depth map on the coarse scale by downsampling the image and then guiding the fine-scale image for depth estimation. In addition to adjusting the image resolution, MARMVS adjusts the scale of the patch by measuring the matching, thus ensuring the validity of the photometric consistency metric.

Another class of approaches uses prior information to obtain reliable depth estimates. TAPA-MVS\cite{TAPA} assumes that untextured areas are piecewise planar, which divides the image into superpixels and then uses an initial depth map to fit a plane to each superpixel for estimating the depth of untextured areas. ACMP\cite{ACMP} samples some reliable pixels and triangulates them to obtain a planar prior model, and it designs a matching cost that considers photometric consistency and planar compatibility to make it applicable to depth estimation in non-planar and planar areas.

untextured areas and non-Lambert surfaces are difficult to estimate robustly by traditional approaches. Therefore, some approaches started to try the combination of PatchMatch with deep learning and achieve good results. The sky areas of outdoor scenes often produce strong artifacts that cannot be filtered by geometric consistency. PCF-MVS \cite{ref28} introduces a semantic segmentation-based sky areas detection to reject artifacts. Depth map confidence estimation is a key component in 3D reconstruction, which optimizes the depth map and filters outliers. DeepC-MVS designed a DNN-based confidence prediction network, which can reliably perform depth map optimization and outlier filtering in high-resolution images. Similarly, \cite{ref10} used point clouds to generate a surface mesh model, which is used to guide the depth estimation of untextured areas, and it designs a neural network module to detect the confidence level of the depth map and reject the points with low confidence. 

In summary, there are two main types of approaches to achieve depth estimation of untextured areas. One category is to increase the receptive field of the patch, and the other category is to use prior information or confidence predictions to optimize the results of the photometric consistency metric. There are limitations in using a single approach, so some approaches \cite{ref10,ref11,ref12} use ACMM as the base module and then optimize the depth map by their own approach. However, these approaches perform their own optimization approach at each scaled image, and this computation increases the running cost and redundancy, and there is an accuracy loss in the upsampling. As with MAR-MVS, we note that expanding the patch size can increase the receptive field. Based on this fact, our idea is to use multi-scale windows in one PatchMatch process to achieve concise and fast multi-scale depth estimation. At the same time, we introduce planar prior assisted PatchMatch and refine it by geometric consistency.

\section{PROPOSED APPROACH}
Given a set of calibration images $I=\left \{ I_{i}|i=1,2,...,N  \right \}$ and a set of corresponding camera parameters $P=\left \{ P_{i}|i=1,2,...,N  \right \}$, MVS needs to estimate the depth maps $D=\left \{ D_{i}|i=1,2,...,N  \right \}$ of all images. The pipeline of our approach is shown in Fig. 1.

\begin{figure*}[!t]\centering\includegraphics[width=18cm]{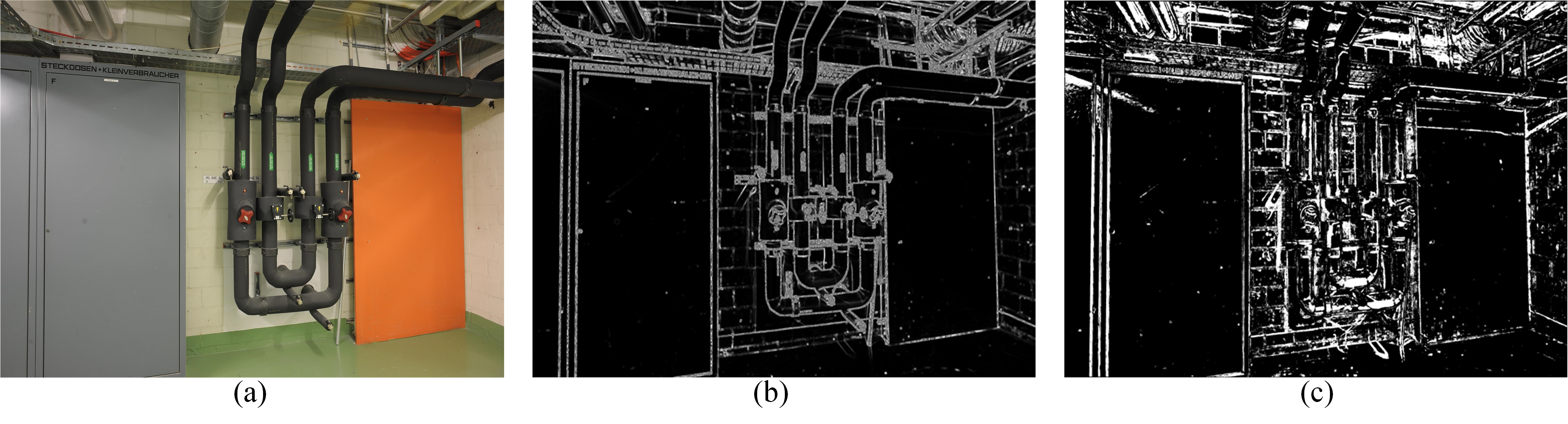}\caption{Image(a) shows the original image. Image(b) shows the visualization of the textureness coefficients. Image (c) shows the binary image of the photometric consistency cost. The photometric consistency cost threshold is 0.1, Areas where the cost is less than 0.1 are shown in white.}\label{fig_3}
\end{figure*}

\begin{figure}[!t]\centering\includegraphics[width=8.5cm]{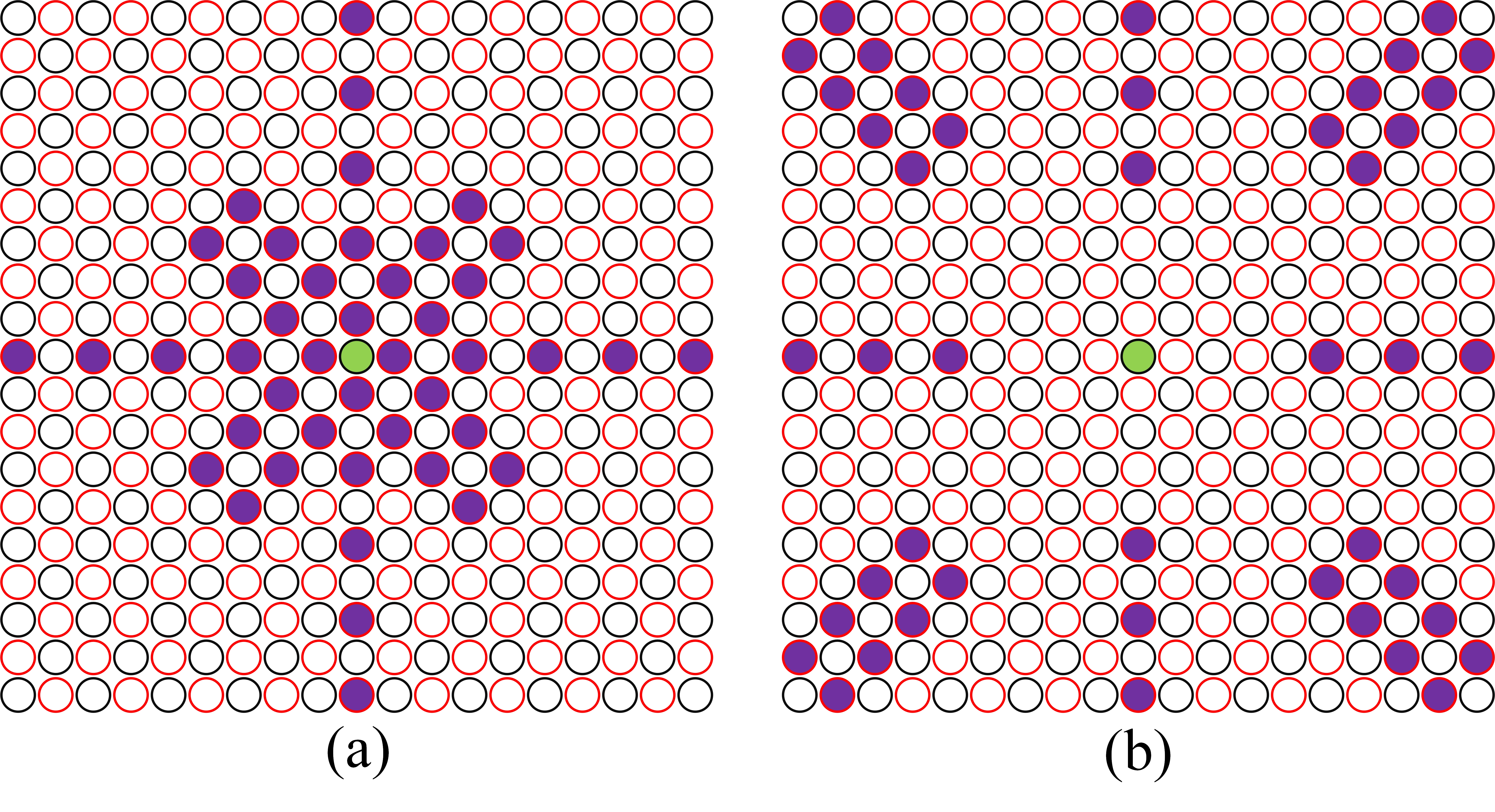}\caption{Image(a) shows ACMH's checkerboard sampling pattern. Image (b) shows our checkerboard sampling pattern.}\label{fig_4}
\end{figure}

During one PatchMatch process, we first estimate the depth using the largest scale window. Then, we progressively decrease the window size and refine the depth estimation results for each pixel through iterative propagation processes. Moreover, we improve the checkerboard sampling scheme of ACMH. By eliminating closer sampling points and only sampling optimal planar hypotheses from the eight more distant neighborhoods, we effectively suppress outlier generation, thereby improving the efficiency and reliability of spatial propagation. Finally, we introduce planar prior assisted PatchMatch of ACMP.  Instead of using photometric consistency as in \cite{ACMP}, we employ geometric consistency to generate the planar prior model to improve the completeness of untextured scene reconstruction.
\subsection{Multi-scale windows PatchMatch}
We start by introducing the basic PatchMatch framework (ACMH). PatchMatch converts the depth estimation into a planar estimation problem, where the planar hypothesis of each pixel in the image can be obtained by random generation. Assuming that the plane hypothesis of pixel $p$ in the reference image is $\theta$, for reference image patch $W_{j}$ which is centered on pixel $p$, we can obtain the corresponding source image patch $W_{i}$ by homography. If the planar hypothesis of pixel $p$ is correct, the texture of corresponding patches should be similar. Then the photometric consistency cost $m_{i,j}$ between corresponding patches can be calculated as
\begin{equation}
m_{i,j}(p,\theta )=\frac{cov(W_{i},W_{j} )}{\sqrt{cov(W_{i},W_{i} )cov(W_{j},W_{j} )} }
\end{equation}
Where $cov(x,y)=E(x-E(x))E(y-E(y))$, and $E(.)$ is the expected value of pixel intensities within a patch.

Define weights with each source image as $\omega_{j}$. The multi-view aggregated photometric consistency cost $m_{p}$ for each pixel with plane hypothesis is defined as
\begin{equation}
m_{p}(p,\theta )=\frac{ {\textstyle \sum_{j=1}^{n}}\omega_{j}\cdot m_{i,j}   }{ {\textstyle \sum_{j=1}^{n}}\omega_{j}} 
\end{equation}

PatchMatch samples the planar hypothesis of neighboring pixels and computes aggregation cost. If the aggregation cost of the plane hypothesis is smaller, it is propagated to the current pixel. The PatchMatch typically performs three iterations of propagation optimization with a fixed window size. Unlike the base PatchMatch, our proposed mPM is propagation optimized with multi-scale windows. To ensure that propagation optimized can be implemented consistently and reliably, The following principles should be observed:

\begin{itemize}
\item{A larger scale window implies more computational cost, and to reduce the computation time, the pixels within the window should be sparsely sampled with equal spacing(Fig. 2), and the photometric consistency metric is computed by sampling points only.}
\item{To ensure that the spatial propagation process is efficient, window size should be the same for all pixels in the same iteration. Similarly, to ensure that all planar hypotheses can be refined at finer scales, the number of sampling points should be the same at different scales.}
\item{Larger-scale windows have a larger receptive field and are therefore able to estimate planar hypotheses for untextured areas more reliably. However, due to the effect of sparse sampling, the depth estimation accuracy obtained from larger-scale windows is poor. Therefore, iterative propagation should be performed starting from a larger scale window and gradually decreasing the window size.}
\end{itemize}

Based on the above criteria, we sample 36 points at equal intervals in a window (Fig. 2). We define the window edge size $L$ at each scale $S$ as
\begin{equation}
L=2^{S}\times 10+1,S\in (0,S_{max})
\end{equation}

The maximum scale of windows is determined according to the resolution of images. Starting from the maximum scale window, three iterations of propagation optimization are executed at each scale, and then the scale is reduced step by step for a total of  $3\times S$ iterations.

Note that at each scale we reset the cost threshold $\tau (t)$ for a good match. To ensure that points that did not obtain reliable planar hypotheses at the previous scale get a robust view selection at this scale. For the t-th iteration under each scale window, the good match cost threshold $\tau (t)$ can be calculated as
\begin{equation}
\tau (t)=\tau_{0}\cdot e^{-\frac{t^{2}}{\alpha } }
\end{equation}
where $\tau_{0}$ is the initialized matching cost threshold and $\alpha$ is a constant.

\begin{figure*}[!t]\centering\includegraphics[width=15cm]{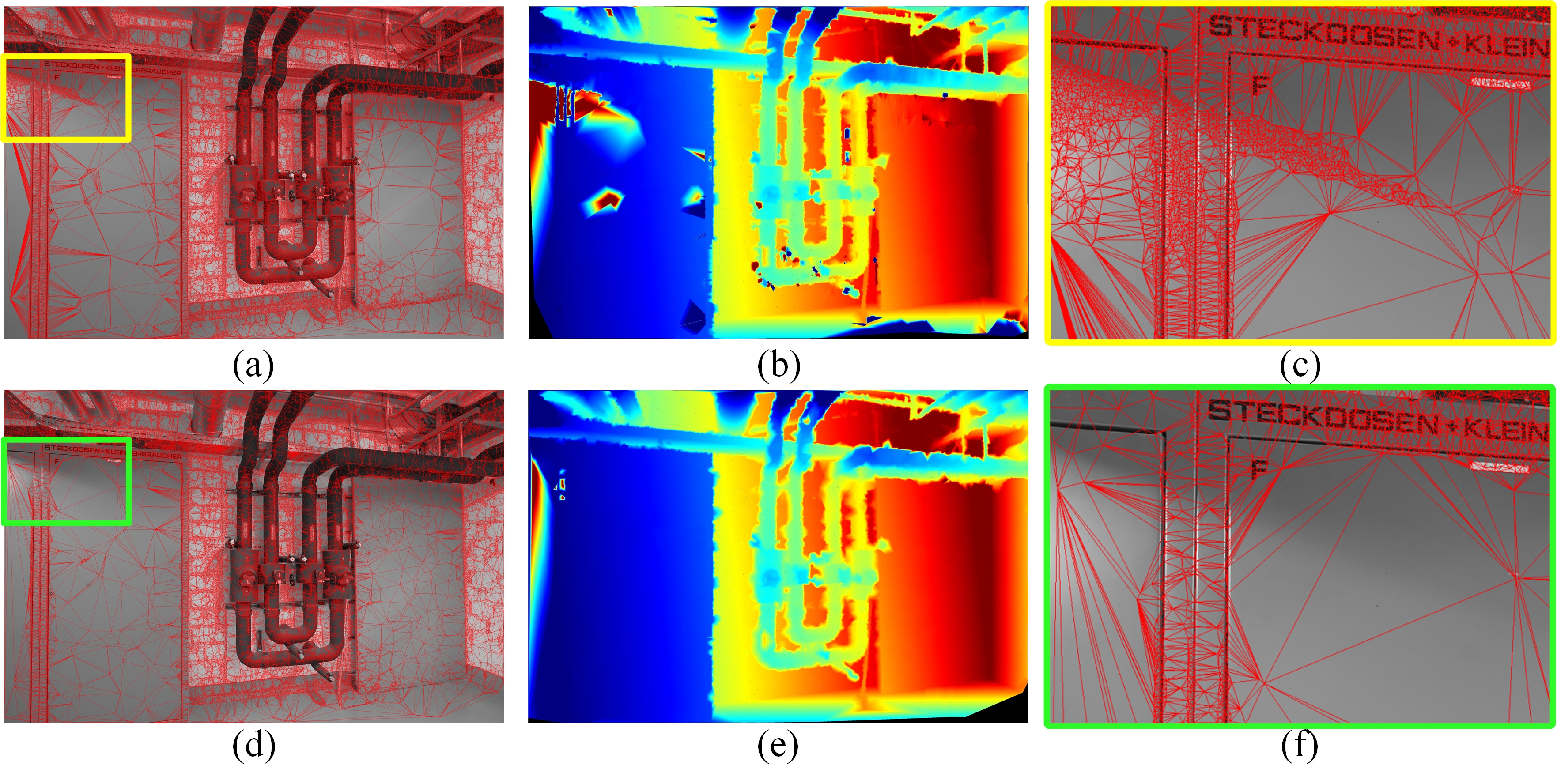}\caption{The effect of ACMP's planar prior model (top) and the effect of our proposed geometric consistency cost to con-struct planar prior model (bottom). The left figures show the effect of triangulation, the middle figures show the depth map generated by the triangulated mesh, and the right figures show the local areas detail of the triangulated.}\label{fig_5}
\end{figure*}

\subsection{Checkerboard sampling}
ACMH\cite{ref9} divides neighborhood pixels into eight regions and adaptively selects the optimal hypotheses of each region to propagate (Fig. 4a). Compared to propagation using only neighboring pixels in \cite{ref23}, this adaptive sampling scheme makes it more likely to sample good hypotheses. Since photometric consistency does not enable reliable depth estimation in untextured areas, only a few reliable planar hypotheses exist for untextured areas (Fig. 3c), and most of the reliable planar hypotheses correspond to texture-rich areas (Fig. 3b). As a result, closer ground sampling in the untextured areas often fails to obtain reliable planar hypotheses, thus hindering effective spatial propagation. This leads to a decrease in the accuracy of the depth estimation. On the other hand, it leads to an increase in the number of outliers. It is more severe by the increase in untextured areas of high-resolution images. 

Therefore, we do not sample planar hypotheses in the region near the estimated pixel, but set the sampling region far from the estimated pixel(Fig. 4b). For the four regions of top, bottom, left, and right, we set 10 samples in each region. In the four diagonal regions, we set 12 samples in each region. Our proposed strategy can effectively improve the reliability of depth estimation in untextured areas while suppressing the generation of outliers. Experiments show that our sampling strategy does not affect the depth estimation of thin objects.

\subsection{Planar prior model}
There are many planar structures in architectural scenes, that can provide effective planar priors information to assist depth estimation, and many Approaches \cite{ACMP,ref10,TAPA,ref30,ref31} carry out research on structure priors.

We introduce the planar prior assisted PatchMatch of \cite{ACMP} to improve the completeness of the reconstruction of untextured areas. ACMP divides a whole image into multiple rectangular regions, and a point with the smallest photometric consistency cost is selected in each rectangular region for triangulation, thus the accuracy of the plane prior relies on the reliability of selected points. Due to the ambiguity of photometric consistency in untextured areas, the depths of some of the selected points are wrong (fig. 5c), which results in generating a wrong planar prior model. Prior to COLMAP's integration of geometric consistency into PatchMatch inference, geometric consistency typically filtered out outliers only in the post-processing stage. A multi-view geometric consistency can effectively refine depth estimation results. We note that using multi-view geometric consistency can find accurate sparse points in textureless regions. A robust strategy for generating planar prior models is to choose triangulated points by geometric consistency. Therefore, instead of directly generating planar prior assisted PatchMatch after computing an initial depth map, we first perform geometric consistency optimization, and the geometric consistency cost $m_{g}$  is calculated as
\begin{equation}
m_{g}(p,\theta )=\frac{ {\textstyle \sum_{j=1}^{n}}\omega_{j}\cdot min(\bigtriangleup e_{i,j},2.0)   }{ {\textstyle \sum_{j=1}^{n}}\omega_{j}} 
\end{equation}
where $\bigtriangleup e_{i,j}$ is the forward-backward reprojection error between the reference image $I_{j}$ and the source image $I_{i}$.

We select points that satisfy $m_{p}<0.2$ and $m_{g}<1.0$ within each region (Fig. 5d). Fig. 5e shows that geometric consistency can eliminate most of the outliers. Compared to ref{ACMP}, we generated a planar prior model that is more accurate.

\begin{figure*}[!t]\centering\includegraphics[width=18cm]{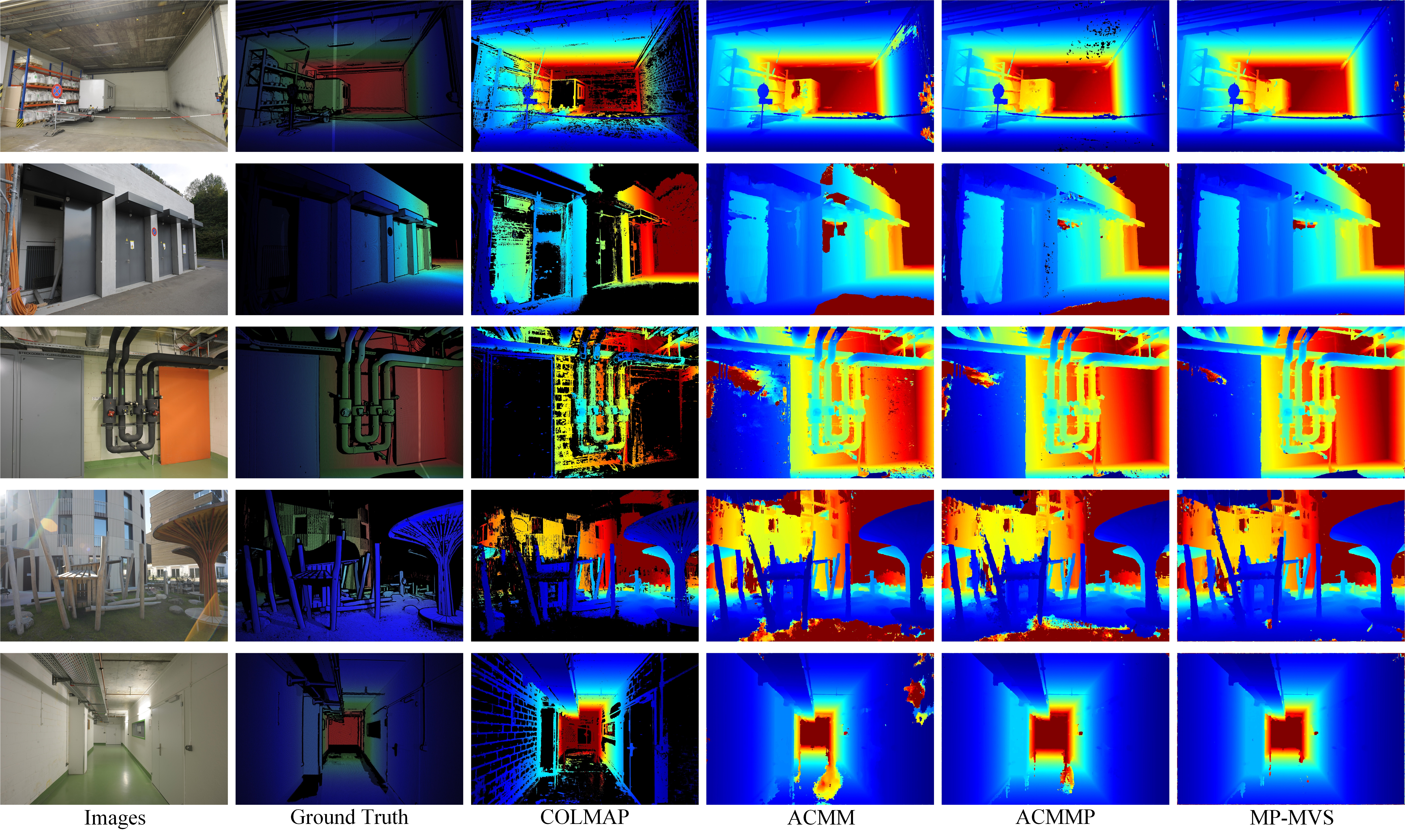}\caption{Qualitative depth map comparison between different approaches on the ETH3D high-resolution dataset (delivery\_area, electro, pipes, playground, terrains). The contrast of the depth map is enhanced by limiting outliers.}\label{fig_6}
\end{figure*}

\begin{table}[!t]
\caption{Results ($F_{1}$-score/ Accuracy/ Completeness) obtained by our approach on the ETH3D high-resolution training dataset with different settings, where rs, pp, and gp stand for our refine checkerboard sampling scheme, planer prior model (ACMP), and our geometric consistency generates plane prior model.\label{tab:table1}}
\centering
\begin{tabular}{c c c c c c c}
\hline
\multirow{3}{*}{Approach} & \multicolumn{6}{c}{Train}                         \\ 
                          & \multicolumn{3}{c}{2cm} & \multicolumn{3}{c}{5cm} \\ \cmidrule(lr){2-4} \cmidrule(lr){5-7}
                          & $F_{1}$      & Acc      & Comp     & $F_{1}$       & Acc     & Comp     \\
\hline
\multicolumn{1}{c}{Baseline(ACMH)}     & 79.55  & 92.04  & 70.74  & 89.87  & 96.80  & 84.33     \\
\multicolumn{1}{c}{mPM+ACMH}      & 82.25  & 91.28  & 75.24  & 91.58  & 96.36  & 87.43     \\
\multicolumn{1}{c}{mPM+rs}        & 83.63  & 91.26  & 77.52  & 92.29  & 96.46  & 88.62     \\
\multicolumn{1}{c}{mPM+ACMH+pp}   & 83.81  & 91.01  & 77.93  & 92.68  & 96.25  & 89.48     \\
\multicolumn{1}{c}{mPM+rs+pp}     & 85.01  & 90.83  & 80.08  & 93.10  & 96.21  & 90.29     \\
\multicolumn{1}{c}{mPM+rs+gp}     & 85.50  & 90.84  & 80.94  & 93.35  & 96.13  & 90.81     \\
\hline
\end{tabular}
\end{table}

\section{Experiment}
To validate the effectiveness of our approach, we evaluated our approach on the Intel core i7 11700 and RTX 3060 against the ETH3D High-res multi-view benchmark. The dataset was divided into a training set and a test set, which included 9 outdoor scenes and 16 indoor scenes. The original resolution of the images was 6048×4032, and in the experiments, we adjusted the image resolution uniformly to 3200×2130.

\begin{figure*}[!t]\centering\includegraphics[width=18cm]{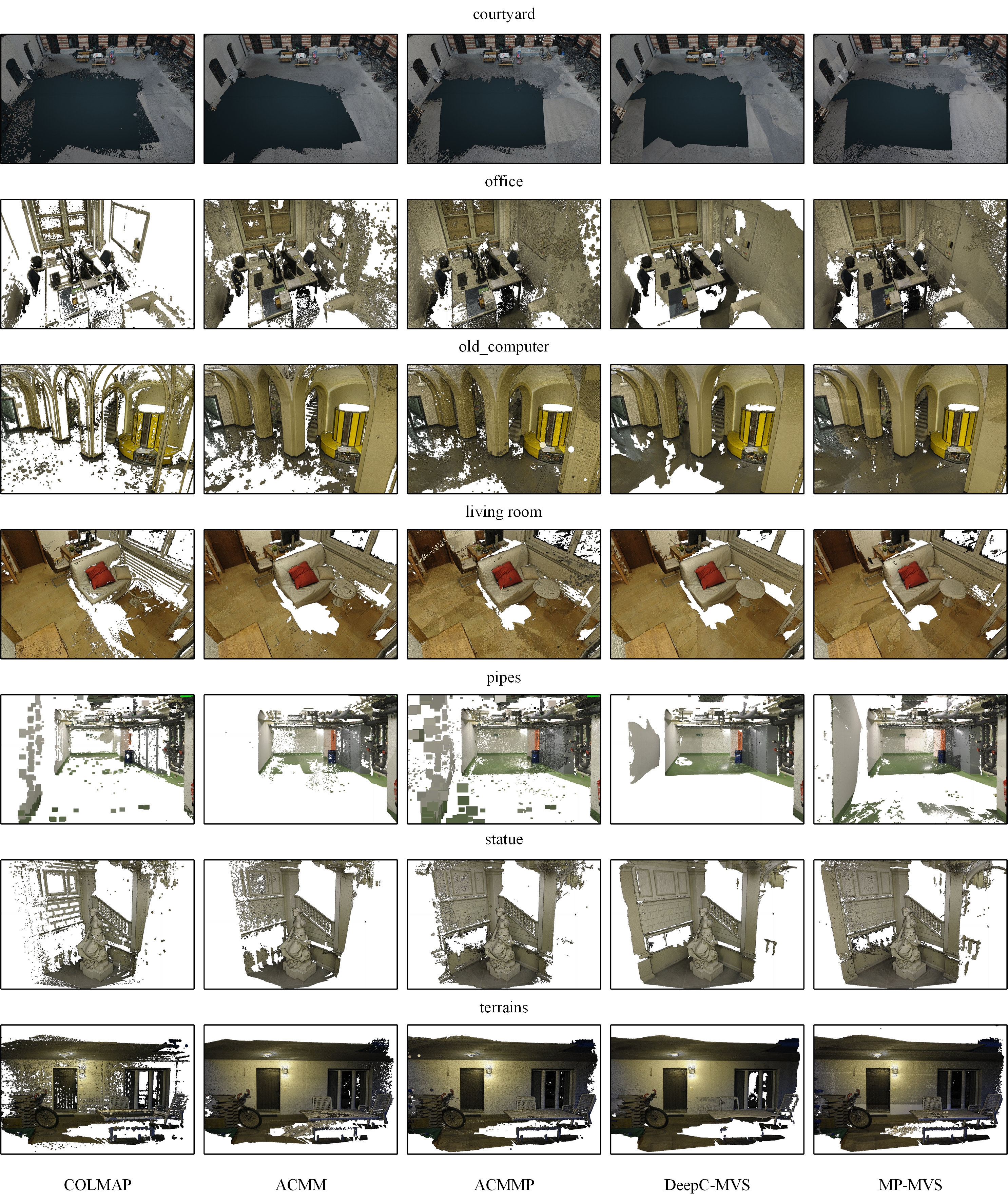}\caption{Comparison of qualitative point cloud reconstruction results between different approaches on the ETH3D high-resolution dataset of dense point cloud models from the ETH3D benchmark server.}\label{fig_7}
\end{figure*}

\begin{table*}[!t]
\caption{Point clouds are evaluated on the training and test sets in the ETH3D high-resolution dataset, with bolded representing the optimal score and red font representing the suboptimal score.\label{tab:table2}}
\centering
\begin{tabular}{ c c c c c c c c c c c c c c}
\hline
\multirow{3}{*}{Scene} & \multirow{3}{*}{Approach} & \multicolumn{6}{c}{Train} & \multicolumn{6}{c}{Test}    \\
                        &                           & \multicolumn{3}{c}{2cm} & \multicolumn{3}{c}{5cm}& \multicolumn{3}{c}{2cm} & \multicolumn{3}{c}{5cm} \\
\cmidrule(lr){3-5} \cmidrule(lr){6-8}\cmidrule(lr){9-11} \cmidrule(lr){12-14}
                        &                           & $F_{1}$   & Acc   & Comp   & $F_{1}$    & Acc   & Comp & $F_{1}$   & Acc   & Comp   & $F_{1}$    & Acc    & Comp    \\
\hline
\multirow{7}{*}{Indoor} &\multicolumn{1}{c}{ACMM}     & 78.13  & 92.46  & 68.49   & 86.09  & 96.42  & 78.39    & 79.84  & 90.99  & 72.73   & 88.48  & 96.07  & 82.85   \\
                        &\multicolumn{1}{c}{MAR-MVS}  & 80.32  & 82.92  & 78.37   & 88.44  & 91.14  & 86.30    & 80.70  & 78.74  & 83.43   & 89.37  & 88.87  & 90.22   \\
                        &\multicolumn{1}{c}{COLMAP}   & 66.76  & \textbf{95.01}  & 52.90   & 78.49  & \textbf{97.98}  & 66.61    
                                                      & 70.41  & \textbf{91.95}  & 59.65   & 82.04  & \textbf{96.62}  & 73.00   \\
                        &\multicolumn{1}{c}{CLD-MVS}  & 81.23  & 87.22  & 77.29   & 88.52  & 94.32  & 84.24    & 81.65  & 82.64  & 82.35   & 88.90  & 91.41  & 87.48   \\
                        &\multicolumn{1}{c}{ACMMP}    & 85.27  & 92.40  & 79.59   & \color{red}{92.16}  & 96.27  & \color{red}{88.71}    
                                                      & 85.39  & \color{red}{91.87}  & 80.67   & 92.88  & \color{red}{96.34}  & 89.97   \\
                        &\multicolumn{1}{c}{DeepC-MVS}& \color{red}{86.03}  & \color{red}{92.56}  & \color{red}{81.05}   & 91.86  & \color{red}{96.75}  & 88.03    
                                                      & \color{red}{86.88}  & 89.08  & \textbf{85.24}  & \color{red}{92.99}  & 95.33  & \color{red}{91.06}   \\
                        &\multicolumn{1}{c}{Ours}     & \textbf{87.40}  & 91.95  & \textbf{83.49}   & \textbf{93.59}  & 95.93  & \textbf{91.45}    
                                                      & \textbf{87.64}  & 90.91  & \color{red}{84.97}   & \textbf{94.07}  & 95.68  & \textbf{92.67}   \\
\hline
\multirow{7}{*}{Outdoor}&\multicolumn{1}{c}{ACMM}     & 79.71  & 88.57  & 72.67   & 89.54  & \color{red}{96.19}  & 83.86    & 83.58  & 89.63  & 79.17   & 91.12  & 96.97  & 86.31   \\
                        &\multicolumn{1}{c}{MAR-MVS}  & 77.92  & 80.88  & 75.81   & 88.33  & 92.71  & 84.88    & 85.27  & 84.73  & \textbf{86.44}   & 93.07  & 94.66  & 91.86   \\
                        &\multicolumn{1}{c}{COLMAP}   & 68.70  & 88.16  & 57.73   & 82.85  & 96.05  & 73.76    & 80.81  & \textbf{92.04}  & 72.98   & 89.74  & 97.13  & 83.94   \\
                        &\multicolumn{1}{c}{CLD-MVS}  & 77.16  & 77.54  & 77.45   & 88.82  & 91.05  & 86.83    & 84.29  & 84.79  & 83.86   & 91.18  & 93.04  & 89.44   \\
                        &\multicolumn{1}{c}{ACMMP}    & 81.26  & \color{red}{88.56}  & 75.30   & 91.87  & 95.93  & 88.22   
                                                      & 87.38  & \color{red}{92.03}  & 83.93   & \color{red}{94.30}  & \color{red}{97.43}  & 91.64   \\
                        &\multicolumn{1}{c}{DeepC-MVS}& \textbf{83.37}  & 87.82  & \textbf{79.42}   & \color{red}{92.73}  & 96.08  & \color{red}{89.71}    
                                                      & \color{red}{87.69}  & 89.37  & \color{red}{86.37}   & 94.26  & 95.71  & \textbf{92.94}   \\
                        &\multicolumn{1}{c}{Ours}     & \color{red}{83.28}  & \textbf{89.55}  & \color{red}{77.96}   & \textbf{93.08}  & \textbf{96.37}  & \textbf{90.06}    
                                                      & \textbf{89.93}  & 91.83  & 85.17   & \textbf{94.77}  & \textbf{97.51}  & \color{red}{92.42}   \\
\hline
\multirow{7}{*}{All}    &\multicolumn{1}{c}{ACMM}     & 78.86  & 90.67  & 70.42   & 87.68  & 96.31  & 80.91    & 80.78  & 90.65  & 74.34   & 89.14  & 96.30  & 83.72   \\
                        &\multicolumn{1}{c}{MAR-MVS}  & 79.21  & 81.98  & 77.19   & 88.39  & 91.86  & 85.65    & 81.84  & 80.24  & 84.18   & 90.30  & 90.32  & 90.63   \\
                        &\multicolumn{1}{c}{COLMAP}   & 67.66  & \textbf{91.85}  & 55.13   & 80.50  & \textbf{97.09}  & 69.91    
                                                      & 73.01  & \textbf{91.97}  & 62.98   & 83.96  & \textbf{96.75}  & 75.74   \\
                        &\multicolumn{1}{c}{CLD-MVS}  & 79.35  & 82.75  & 77.36   & 88.66  & 92.81  & 85.43    & 82.31  & 83.18  & 82.73   & 89.47  & 91.82  & 87.97   \\
                        &\multicolumn{1}{c}{ACMMP}    & 83.42  & 90.63  & 77.61   & 92.03  & 96.12  & 88.48    
                                                      & 85.89  & \color{red}{91.91}  & 81.49   & 93.24  & \color{red}{96.61}  & 90.39   \\
                        &\multicolumn{1}{c}{DeepC-MVS}& \color{red}{84.81}  & 90.37  & \color{red}{80.30}   & \color{red}{92.26}  & \color{red}{96.44}  & \color{red}{88.80}    
                                                      & \color{red}{87.07}  & 89.15  & \textbf{85.52}   & \color{red}{93.31}  & 95.43  & \color{red}{91.53}   \\
                        &\multicolumn{1}{c}{Ours}     & \textbf{85.50}  & \color{red}{90.84}  & \textbf{80.94}   & \textbf{93.35}  & 96.13  & \textbf{90.81}    
                                                      & \textbf{87.71}  & 91.14  & \color{red}{85.02}   & \textbf{94.25}  & 96.14  & \textbf{92.61}   \\
\hline
\end{tabular}
\end{table*}

\begin{table*}[!t]
\caption{Average running time (per second) for depth map estimation by different approaches on the ETH3D dataset.\label{tab:table3}}
\centering
\begin{tabular}{c c c c c c c c c c c c c c c}
\hline
Approach	&Ave.	&counry.	&deli.	&elec.	&faca.	&kick.	&mead.	&offi.	&pipes	&playgr.	&relif	&relif2	&terra.	&terrain.\\
\hline
COLMAP    &76.67	&74.09	&75.25	&79.38	&81.65	&67.32	&61.65	&85.57	&69.98	&71.26	&83.49	&63.91	&75.01	&87.71 \\
ACMMP     &35.49	&35.06	&30.69	&33.12	&42.03	&39.37	&31.53	&29.69	&23.39	&36.37	&38.86	&35.48	&30.04	&37.53 \\
ACMM      &\textbf{28.70}	&28.17	&24.78	&\textbf{26.74}	&\textbf{35.94}	&\textbf{31.89}	&\textbf{25.08}	&\textbf{24.21}	&18.82	&28.14	&\textbf{30.14}	&\textbf{28.07}	&\textbf{24.39}	&\textbf{29.68} \\
Ours    &30.32	&\textbf{27.76}	&\textbf{24.36}	&27.16	&36.96	&37.98	&28.70	&25.17	&\textbf{18.29}	&\textbf{27.68}	&34.94	&30.28	&24.82	&34.29 \\
\hline
\end{tabular}
\end{table*}

\subsection{Ablation study}
We performed ablation experiments on the training set of ETH3D for validating the effects of our mPM, checkerboard sampling scheme, and geometric consistency generating plane prior model. We validate the effectiveness of our approach by using ACMH as our baseline method and adding our individual modules incrementally. For depth map fusion, we used the latest version scheme of ACMMP to ensure that the best results are obtained for each module. The results are shown in Table 1. 

Compared to baseline, mPM significantly improves the completeness of scene reconstruction through multi-scale windows, while this coarse-to-fine approach has less impact on the loss of accuracy. Similarly, our proposed checkerboard sampling scheme is able to propagate correct hypotheses more efficiently in untextured regions, which is why the completeness of scene reconstruction is significantly improved. Generating a planar prior model from geometric consistency information can obtain more accurate results than using photometric consistency directly. The above experiments show that our proposed approach can well balance accuracy and completeness, and can significantly improve depth estimation of untextured regions with a small loss of accuracy.

\subsection{Evaluation}
We select some SOTA approaches for comparison with ours, and in Table 2 we list the training and test set results of these approaches for the ETH3D high-resolution multi-view benchmark, which are publicly available on the ETH3D benchmark server. Indoor scenes often have a large number of artificial objects, which results in more untextured areas. Since we use several methods to improve the completeness of untextured areas, our approach overall obtains optimal completeness and $F_{1}$-score in indoor scenes. In more complex outdoor scenes, our proposed approach also shows strong performance in terms of accuracy. Overall, our approach is highly adaptable to both indoor and outdoor scenes.

In addition to quantitative analysis, we have also selected some open-source approaches for visual comparison to show the performance of our approach more intuitively. We selected five scenes for depth map estimation: delivery\_area, electro, pipes, playground, and terrains. The results of the depth map visualization obtained by several approaches are presented in Fig. 6. COLMAP eliminates unreliable depth estimates by a strict strategy but at the expense of depth map completeness. ACMM and ACMMP obtain good results in untextured areas, but in some challenging areas that are only observed by a few images, these approaches have difficulty in estimating correct depth. It can be seen that the ground in part of the scenes (electro, playground, terrains) only has the correct depth calculated by ours. The playground scene has many thin foregrounds, and our depth estimation results are significantly better than other approaches for thin foregrounds,  which confirms that our proposed checkerboard sampling scheme has no significant effect on thin objects.

Comparing point cloud reconstruction results with those publicly available on the ETH3D benchmark server (Fig. 7). From point cloud reconstruction results, our approach can obtain the most complete results, especially for some untextured areas, and our approach is able to obtain more robust results.

\subsection{Runtime performance}
We list some approaches and ours to estimate the average running time (per second) of each depth map in Table 3. Considering the variability of data sizes and scenarios, we detail runtimes for all scenes in the ETH3D training dataset, and all approaches running on our platform based on a single GPU. Where ACMM and ACMMP downsample all images 2 times, ACMM performs 7 iterations and ACMMP performs 10 iterations in each scale image. COLMAP performs 5 iterations in a single scale image and our approach performs 16 iterations in a fixed scale image.

Since our mPM simplifies the multi-scale depth estimation, our approach is close to ACMM in terms of running time in most scenes, and even faster in some scenes. Our approach is 17\% faster than ACMMP which also uses a multi-scale approach and planar prior assisted PatchMatch, while the performance of our approach is stronger than the above approaches.

\section{Conclusion}
In this paper, we propose a multi-scale windows Patch-Match (mPM) that can be easily extended to PatchMatch-based MVS approaches. In addition, we optimize the existing checkerboard sampling scheme for more effective spatial propagation, and we generate more accurate planar prior model by geometric consistency. These strategies effectively improve the completeness of scene reconstruction. We evaluated our approach on the ETH3D dataset, which integrates the above strategies. The results show that our method enables state-of-the-art 3D reconstruction in high-resolution datasets.


\vfill

\end{document}